\pdfoutput=1

\documentclass[11pt]{article}

\usepackage[final]{acl}
\usepackage{array}
\usepackage{times}
\usepackage{latexsym}
\usepackage{amsmath}
\usepackage{kotex}
\usepackage{tabularx} 
\usepackage{makecell}
\usepackage{amssymb}
\usepackage{pifont}
\usepackage{xcolor}
\usepackage[T1]{fontenc}

\usepackage[utf8]{inputenc}

\usepackage[protrusion=false, expansion=false]{microtype}
\usepackage{inconsolata}

\usepackage{graphicx}
\usepackage{longtable}
\usepackage{listings}
\usepackage{tcolorbox}
\usepackage{tikz}
\tcbuselibrary{skins}
\usepackage{setspace}
\usepackage{amsmath}
\usepackage{booktabs}    
\usepackage{multirow} 

\definecolor{DarkGreen}{RGB}{0,225,0}
\definecolor{DarkRed}{RGB}{255,0,0}
\definecolor{slategray}{RGB}{112, 128, 144}
\definecolor{teal}{RGB}{0, 128, 128}
\definecolor{royalblue}{RGB}{65, 105, 225}
\definecolor{darkolivegreen}{RGB}{85, 107, 47}
\definecolor{rosegold}{RGB}{183, 110, 121}
\definecolor{midnightblue}{RGB}{30, 30, 122}
\definecolor{lavender}{RGB}{150, 123, 182}
\definecolor{goldenrod}{RGB}{218, 165, 32}
\definecolor{darkgoldenrod}{RGB}{184, 134, 11}
\definecolor{slateblue}{RGB}{106, 90, 205}
\definecolor{dimgray}{RGB}{100, 100, 140}
\usepackage{algorithm}
\usepackage{algorithmic}

\usepackage{authblk}

\title{FeRG-LLM : Feature Engineering by \\ Reason Generation Large Language Models}

\author{\textbf{Jeonghyun Ko}\textsuperscript{1}\thanks{equal contribution.}, \textbf{Gyeongyun Park}\textsuperscript{1*}, \textbf{Donghoon Lee}\textsuperscript{1*}, \textbf{Kyunam Lee}\textsuperscript{2*}\thanks{corresponding author.}}

\affil{Korea University\textsuperscript{1},\quad SK Telecom \textsuperscript{2}} \affil{\texttt{\{2014jhyun, pk53ar, ydh82066\}@gmail.com, kyunam@sk.com}}

\date{}

\begin{document}
\maketitle

\begin{abstract}
One of the key tasks in machine learning for tabular data is feature engineering. Although it is vital for improving the performance of models, it demands considerable human expertise and deep domain knowledge, making it labor-intensive endeavor. To address this issue, we propose a novel framework, \textbf{FeRG-LLM} (\textbf{Fe}ature engineering by \textbf{R}eason \textbf{G}eneration \textbf{L}arge \textbf{L}anguage \textbf{M}odels), a large language model designed to automatically perform feature engineering at an 8-billion-parameter scale. We have constructed two-stage conversational dialogues that enable language models to analyze machine learning tasks and discovering new features, exhibiting their Chain-of-Thought (CoT) capabilities. We use these dialogues to fine-tune Llama 3.1 8B model and integrate Direct Preference Optimization (DPO) to receive feedback improving quality of new features and the model's performance.
Our experiments show that FeRG-LLM performs comparably to or better than Llama 3.1 70B on most datasets, while using fewer resources and achieving reduced inference time. It outperforms other studies in classification tasks and performs well in regression tasks. Moreover, since it does not rely on cloud-hosted LLMs like GPT-4 with extra API costs when generating features, it can be deployed locally, addressing security concerns.
\end{abstract}

\begin{figure*}[!h]
    \centering
    \includegraphics[width=1\linewidth]{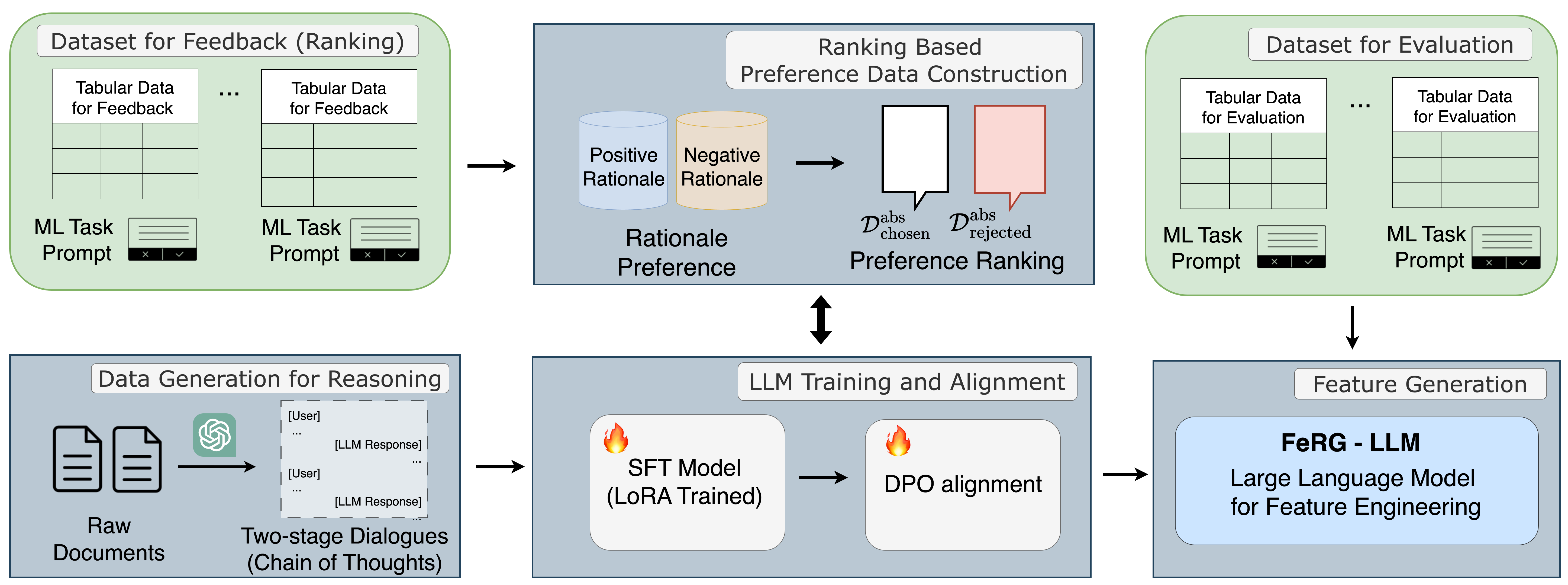}
    \caption{Overall framework of FeRG-LLM. The method first generates a two-stage dialogue related to feature discovery and then performs SFT using LoRA. It then optimizes the language model to provide reasoning feedback.}
    \label{fig:overall-workflow}
\end{figure*}

\section{Introduction}
\hspace{15pt} While deep learning has shown remarkable performance on diverse structure of datasets and gained significant attention, classical machine learning models continue to play a pivotal role across various industries to satisfy interpretability requirements and accommodate computational constraints. These models help organizations make data-driven decisions and build a specialized database of newly discovered key features that significantly enhance model performance. However, uncovering these features requires considerable expertise and analytical skills on the part of data scientists, creating a strong demand for highly qualified professionals.\par
Since recent large language models (LLM) have achieved superior performance \cite{achiam2023gpt, dubey2024llama} in various evaluation metrics due to their extensive knowledge and advanced text-understanding capabilities, several studies have applied language models to feature engineering \cite{hollmann2024large, han2024large}. They have successfully proposed methodologies for classification tasks, but they rely on cloud-hosted LLMs such as OpenAI GPT \cite{achiam2023gpt} so that it cannot be used for sensitive data and remain limited to classification tasks.\par
Therefore, we propose \textbf{FeRG-LLM}, a novel framework that optimizes a language model to discover valuable features for machine learning models. We aim for the language model to fully understand machine learning tasks related to tabular data and generate meaningful features that improve performance. To this end, we construct a dialogue dataset that thoroughly captures the task requirements and yields meaningful features. In the next stage, we train the 8B-scale Llama 3.1 model \cite{dubey2024llama} on these dialogues and then aligned our model using Direct Preference Optimization (DPO) to linguistically generate a more refined rationale for feature creation. Figure \ref{fig:overall-workflow} presents the overall framework.\par
Considering that binary classification tasks are critical in industry, we extensively evaluate our model on 14 relevant datasets. Despite being 8B-scale, the results show that our model achieves performance comparable to that of a 70B model and outperforms other recent studies. In addition to experiments on classification tasks, we demonstrate that our model can be readily applied to other problem types, such as regression, using four additional datasets.

Our contributions are threefold:
 
\begin{itemize}
    \item We propose a novel framework, \textbf{FeRG-LLM}, that harnesses chain-of-thought reasoning through a two-stage dialogue process in a large language model and integrates feedback via DPO to better understand machine learning tasks on tabular data so that valuable features are discovered.
    
    \item Our framework uses Llama 3.1 8B which can be deployed on a single GPU so that it can be used in local environments. Nonetheless, it matches or sometimes even exceeds the capabilities of a 70B model and related work. It greatly reduces resource requirements and eliminates data security concerns making it highly suitable for enterprise adoption.
    
    \item Our method produces executable Python code for feature generation, eliminating manual effort and identifying overlooked patterns. By leveraging only variable names and descriptions, it formulates features without distribution analysis. This automation accelerates data-driven decisions while boosting overall efficiency.\par
\end{itemize}

\section{Related Work}
\paragraph{Feature Engineering with Language Models}
There have been studies of LLMs in the context of feature generation problems for tabular datasets. For instance, CAAFE \cite{hollmann2024large} provides OpenAI GPT with information about a specific Machine Learning (ML) problem including dataset information and prompt it to suggest new features, along with Python code and rationales. It also incorporates feature selection based on features proposed by GPT. However, this approach focuses on small-scale datasets and requires including variable descriptions, specific values, and details about missing data in the prompt. This approach makes working with large-scale datasets difficult and handling security issues challenging, limiting its applicability in corporate or academic research.\newline
Another study by \citet{han2024large} introduces FeatLLM which creates rules to generate binary features from given data using a few-shot approach. It provides specific data values and corresponding labels to OpenAI GPT to generate binary feature representations for classification. Generated features are used in a linear layer to build a classification model. However, its generation of only binary features may limit the diversity of feature engineering and offer limited linguistic interpretability.\newline
We compare the capabilities of FeRG-LLM with these models as outlined in Table \ref{tab:compare_table} and found that it exhibits several advantages over other models.

\begin{table*}[!h]
\centering
\small
\renewcommand{\arraystretch}{1.5} 
\begin{tabular}{p{10.5cm}|cc|c}
\Xhline{1pt}
Criteria  & CAAFE & FeatLLM & \textbf{FeRG-LLM} \\
\hline
\hline
Create contextual features by understanding the dataset and modeling objectives? & \ding{51} & \ding{51} & \ding{51} \\
\hline
Operate without explicitly specifying operators such as inequalities in the prompt?& \ding{51} & \ding{55} & \ding{51} \\
\hline
Applicable to tasks beyond classification (e.g., regression)? & \ding{55} & \ding{55} & \ding{51} \\
\hline
Produce features at the point of generation without external API like GPT-4?& \ding{55} & \ding{55} & \ding{51} \\
\hline
Generate features without requiring specific variable values in the prompt? & \ding{55} & \ding{55} & \ding{51} \\

\Xhline{1pt}
\end{tabular}
\caption{The capabilities of FeRG-LLM are compared with those of other recent models including CAAFE \cite{hollmann2024large} and FeatLLM \cite{han2024large}. A \ding{51} indicates that a criterion is met while a \ding{55} indicates that it is not.
}
\label{tab:compare_table}
\end{table*}

\paragraph{Chain-of-Thought}
Guiding models to solve complex problems through step-by-step reasoning similar to human thinking is referred to as Chain-of-Thought (CoT), and it has become a significant area of research in the field of LLMs. The study \cite{wei2022chain} shows that large language models possess exceptional problem-solving abilities on complex tasks by generating intermediate reasoning steps in text, rather than solely predicting the final answer.\newline
 Meanwhile, \citet{magister2022teaching} notes that CoT is less effective for smaller models. They demonstrate that training smaller models on reasoning patterns from larger models can partially succeed in reproducing CoT. However, they emphasize that this approach relies on the performance of teacher model, making it challenging to enhance the reasoning capability of smaller models.\newline
Nevertheless this study provides insights into the potential for achieving CoT capabilities even in smaller models by focusing on specific tasks. Building on these insights, we employ a two-stage dialogue framework to achieve CoT abilities. This approach fosters a robust understanding of complex machine learning problems and helps uncover meaningful features.

\paragraph{Direct Preference Optimization}

\citet{rafailov2024direct} introduces Direct Preference Optimization (DPO) a method for aligning language models by utilizing human preference data. In contrast the approach of training language models using the Reinforcement Learning from Human Feedback (RLHF) method as proposed in \citet{bai2022training, ouyang2022training} requires training a reward model before optimizing a language model using the PPO algorithm \cite{schulman2017proximal}.
DPO directly optimizes language models by leveraging preference data, eliminating the need to train a reward model. The preference dataset $
\mathcal{D} = \{(x^{(i)}, y^{(i)}_l, y^{(i)}_w)\}_{i=1}^N$ is collected from human annotators who rank responses by preference, where \(y_l\) represents the less preferred response and \(y_w\) the more preferred one.\newline
Studies have extensively investigated the implementation of DPO in LLMs. \citet{allam-2024-biasdpo} proposes BiasDPO, designing DPO training to prioritize the generation of unbiased text on sensitive topics such as gender, race, and religion. They demonstrate that this approach can effectively and reliably mitigate bias. \citet{yang2024direct} demonstrates that by assigning different weights to responses based on the strength of preferences, it is possible to generate responses that better reflect human preferences. \citet{zeng2024token} improves the accuracy of language models in writing and language generation by optimizing at the individual token level, as opposed to the traditional methods that optimize at the sentence or task level.\newline
For our study, we apply DPO instead of RLHF, as it enables automatic feedback on generated rationale in resource-constrained settings. This ensures stable feature generation, boosting machine learning performance.

\section{Methodology}

In this section, we present our FeRG-LLM framework. Feature engineering is not about finding a single correct solution. It requires deep data understanding and the ability to derive meaningful insights through ML and domain expertise. To equip LLMs with such capabilities, we construct a two-stage dialogue dataset for supervised fine-tuning and utilize DPO to offer feedback on the rationale behind feature generation. Furthermore, our framework is designed to efficiently handle large-scale datasets with numerous features and samples, ensuring flexibility and scalability. For space constraints, we provide detailed SFT and DPO training setups, including learning rate, configuration, and other settings, in Appendix \ref{sec:trainsetup}.

\subsection{Two-stage Dialogue for Feature Engineering}

\paragraph{Two-stage Dialogue Generation}

Two-turn dialogue dataset is constructed to leverage the language model’s CoT reasoning capabilities using about 7,190 documents, including papers and materials with data analysis or modeling. GPT-4o-mini API is employed to extract the ML-relevant attributes and contextual task information from each document which is then used to construct a conversational dataset. \newline
The extracted information involves identifying the data domain, describing the variables, clarifying the machine learning problem and its type such as regression or classification, and detailing any innovative features. The specific API prompts used are provided in Appendix \ref{sec:appendixA}.

\begin{figure}[t]
\centering
\begin{tcolorbox}[enhanced, frame style={draw=black, dash pattern=on 3.2pt off 1.6pt}, colframe=white, colback=white, boxrule=0.96pt] 
\spaceskip=0.16em
\setstretch{0.64} 
\footnotesize \ttfamily 

{\footnotesize\textbf{[USER]}}

\vspace{2.4pt} 

\textcolor{darkgoldenrod}{domain : < domain information >}

\vspace{1.6pt} 
\textcolor{darkgoldenrod}{type: < classification/regression/etc >}

\vspace{1.6pt}
\textcolor{darkgoldenrod}{present variable in dataframe:} 

\textcolor{darkgoldenrod}{<feature1 name> : <feature1 description>}

\textcolor{darkgoldenrod}{<feature2 name> : <feature2 description>}

\vdots

\vspace{6.4pt} 

Provide key ideas for creating new derived variables to improve the performance of the following ML problem using the given dataset.

\vspace{6.4pt}

\textcolor{teal}{Machine Learning PROBLEM: <purpose of data analysis>}

\vspace{9.6pt} 

\begin{flushright}
{\footnotesize\textbf{[LLM Assistant]}}

\vspace{2.4pt} 
\textcolor{midnightblue}{1. definition: 1st new feature definition...}

\textcolor{midnightblue}{- 1st new feature description...}

\vspace{6.4pt} 
\textcolor{midnightblue}{2. definition: 2nd new feature definition...}

\textcolor{midnightblue}{- 2nd new feature description...}
\end{flushright}

\hspace{4cm}\textcolor{midnightblue}{\vdots} 

\vspace{9.6pt} 
{\footnotesize\textbf{[USER]}}

\vspace{2.4pt}

Create new features in Python code to enhance machine learning performance using the ideas you provided.

\vspace{9.6pt}
\begin{flushright}
{\footnotesize\textbf{[LLM Assistant]}}

\vspace{1.6pt} 
\textcolor{midnightblue}{\# 1. 1st new feature.}

\vspace{1.6pt}
\textcolor{midnightblue}{df[`1st new'] = ...}

\vspace{6.4pt}
\textcolor{midnightblue}{\# 2. 2nd new feature.}

\vspace{1.6pt}
\textcolor{midnightblue}{df[`2nd new'] = ...}
\end{flushright}

\hspace{4cm}\textcolor{midnightblue}{\vdots} 

\end{tcolorbox}

\caption{Example of two-stage dialogue that facilitates the LLM's reasoning process. The first step involves conceptualizing core ideas, followed by the second step, where these ideas are actualized through the creation of Python code.}
\label{fig:template}
\end{figure}

\paragraph{Reasoning Process}
We reformat the API-extracted information into a two-step dialogue structure to effectively leverage CoT reasoning ability in 8B-scale LLM. In the first turn, the user provides overall information for feature engineering — domain and variables of dataset, the analysis objective, and the ML problem being addressed. At the end of the prompt, the user explicitly instructs the model to provide key ideas for deriving new features. Then the LLM generates the fundamental ideas for creating new features. It is noteworthy that the prompt presents dataset information solely as variable names and descriptions, thereby relieving data analysts from the need to manually examine distributions of each variable.\newline
In the second turn, the model generates Python code to create new features based on the key ideas proposed in its first response. The specific dialogue prompts are shown in Figure \ref{fig:template}. It should be noted that by directly outputting Python code, feature engineering and evaluation can be fully automated. Finally, we carry out supervised fine-tuning methodology LoRA \cite{hu2021lora}, obtaining the fine-tuned model $\pi_{\small{\rm{SFT}}}$.

\subsection{Aligning Language Model with DPO}
\paragraph{Preference Feedback}
To ensure FeRG-LLM generates stable and high-quality features, we perform feature engineering with $\pi_{\rm{SFT}}$ on eight tabular datasets designed for binary classification. These datasets are selected to reflect diverse aspects of ML tasks, forming a robust evaluation set. Detailed information is provided in Appendix \ref{sec:appendixD}.
As is common among data scientists, we use Area Under the Curve (AUC) as the reference metric. Based on this, we provide feedback on the generated rationales.

\paragraph{Preference Data Composition}
For each dataset, we begin with a prompt \( x \) as shown in Figure \ref{fig:template} that contains comprehensive information about the machine learning task. Rationale responses \( R \) are sampled from \( \pi_{\mathrm{SFT}}(\, \cdot | x) \). The language model subsequently produces final features \( F \), and we compute the AUC \( Z \) using XGBoost \cite{Chen:2016:XST:2939672.2939785} when the features \( F \) are included. During inference, we set the temperature to 1.4 and top-p to 0.9. We set this higher temperature to encourage broader feature exploration while ensuring that the model consistently provides rationales and Python code as intended.
This process is repeated for 30 iterations, yielding the sets of rationales \(\{R_i\}_{i=1}^{30}\) and features \(\{F_i\}_{i=1}^{30}\). In fact, we use all combinations of the generated rationales to create preference data, as all cases are explained in the following description.\newline
For each iteration, we compare the AUC value \( Z \) obtained with the derived feature to the baseline $\text{AUC}_{\text{base}}$ from fitting XGBoost to the original dataset. If $Z$ is greater than $\text{AUC}_{\text{base}}$, the corresponding output is stored in 
\[
\mathcal{D}_{\rm{dataset}}^{\rm{pos}} = \{(R_1^{\rm{pos}}, F_1^{\rm{pos}}, Z_1^{\rm{pos}}), \dots\}
\]
otherwise, it is stored in
\[
\mathcal{D}_{\rm{dataset}}^{\rm{neg}} = \{(R_1^{\rm{neg}}, F_1^{\rm{neg}}, Z_1^{\rm{neg}}), \dots\}
\]
We consider two specific criteria: absolute and relative preferences.
\begin{itemize} 
\item \textbf{Absolute preference} Within each dataset, we form all possible combinations of positive and negative rationale pairs using instances from $\mathcal{D}_{\rm{dataset}}^{\rm{pos}}$ and $\mathcal{D}_{\rm{dataset}}^{\rm{neg}}$. In each pair, the $R^{\rm{pos}}$ included in $\mathcal{D}_{\rm{dataset}}^{\rm{pos}}$ is labeled as "chosen," while the $R^{\rm{neg}}$ included in $\mathcal{D}_{\rm{dataset}}^{\rm{neg}}$ is labeled as "rejected."

\item \textbf{Relative Preference} We form pairs of rationales exclusively within each of $\mathcal{D}_{\rm{dataset}}^{\rm{pos}}$ and $\mathcal{D}_{\rm{dataset}}^{\rm{neg}}$. For each pair we compare their AUC scores $Z$. The rationale with the higher score is labeled "chosen" and the other "rejected."
\end{itemize}

Since every possible combination of rationales generated by the LLM falls under one of the two criteria, we construct the preference dataset using all pairs. Finally, we define the preference dataset as $\mathcal{D} = \{(x^{(i)}, r^{(i)+}, r^{(i)-})\}_{i=1}^{M}$, where \( r^+ \) represents the chosen rationale, \( r^- \) represents the rejected rationale and $M$ is the number of all possible combinations. We optimize the language model \(\pi_{\theta}\) to minimize the objective from \citet{rafailov2024direct}:

{
\small
\[
\begin{array}{l}
\mathcal{L}_{\rm{DPO}}(\pi_{\theta}; \pi_{\rm {SFT}}) \\
= - \mathbb{E}_{(x, r^{+}, r^{-}) \sim \mathcal{D}} \left[ \log \sigma\Bigl( p(r^{+} \mid x) - p(r^{-} \mid x) \Bigr) \right]
\end{array}
\]
}
where
\(
p(\cdot| x) = \beta\log\frac{\pi_\theta(\; \cdot \;|\; x)}{\pi_{\rm{SFT}}(\; \cdot \;|\; x)} 
\)
and \(\sigma(\cdot)\) is a logistic function. The parameter \(\beta\), which we set to 0.1 during training, controls the deviation from \(\pi_{\rm SFT}\).

\begin{table*}[t]
\centering
\small
\renewcommand{\arraystretch}{1.2}

\begin{tabular}{cccccccc}
\toprule

\multicolumn{1}{c}{} 
& \multicolumn{7}{c}{\textbf{Dataset}} \\
\cmidrule(lr){2-8}

\textbf{Method} & \textbf{Bank} & \textbf{Nhanes} & \textbf{Diabetes} 
& \textbf{Cultivar} & \textbf{Credit} & \textbf{Ethereum} & \textbf{Churn}\\
\midrule
XGB           
& 93.19          & 75.08          & 79.28          
& 70.64          & 81.65          & 87.91          & 92.12 \\
CAAFE         
& 93.21$\pm$0.32 & 74.47$\pm$0.43 & 81.29$\pm$1.28 
& 71.27$\pm$2.28 & 82.67$\pm$1.65 & 87.94$\pm$0.17 & 92.45$\pm$0.51 \\
FeatLLM 
& 80.05$\pm$0.62 & 66.57$\pm$0.71 & 78.66$\pm$0.47 & 69.52$\pm$2.23 & 50.64$\pm$7.19 & 91.22$\pm$0.73 & 79.91$\pm$0.79 \\
Llama 3.1 70B
& 93.36$\pm$0.05 & \textbf{76.30$\pm$0.42} & \textbf{83.12$\pm$0.38} 
& 76.38$\pm$0.38 & 84.82$\pm$0.27 & \textbf{97.43$\pm$0.29} & \textbf{93.25$\pm$0.27} \\
\textbf{FeRG-LLM}      
& \textbf{93.88$\pm$0.51} & 76.23$\pm$0.40    & 82.43$\pm$0.34 
& \textbf{78.34$\pm$2.74} & \textbf{85.07$\pm$0.67} & 97.03$\pm$0.60 & 93.22$\pm$0.14 \\
\hline
\hline

\multicolumn{1}{c}{} 
& \multicolumn{7}{c}{\textbf{Dataset}} \\
\cmidrule(lr){2-8}

\textbf{Method} & \textbf{Ad-click} & \textbf{Aging-npha} & \textbf{Infrared-t} 
& \textbf{Ilpd-indian} & \textbf{Support2} 
& \textbf{Heart-failure} & \textbf{Glioma} \\
\midrule

XGB           
& 63.85          & 56.97          & 85.74          
& 79.48          & 85.28          & 88.57          & 91.13 \\
CAAFE         
& 65.41$\pm$0.65 & 58.42$\pm$1.94 & 86.30$\pm$0.44 
& 79.48$\pm$0.00 & 85.42$\pm$0.05 & 87.67$\pm$0.62 & 91.11$\pm$0.25 \\
FeatLLM 
& 51.18$\pm$0.57 & 59.16$\pm$0.75 & 80.48$\pm$0.53 & 82.61$\pm$1.17 & 56.90$\pm$7.96 & 80.77$\pm$1.08 & 73.51$\pm$3.28 \\
Llama 3.1 70B
& 70.67$\pm$0.61 & \textbf{61.47$\pm$0.83} & \textbf{87.49$\pm$0.28} 
& 82.94$\pm$0.94 & 85.72$\pm$0.06 & 89.66$\pm$0.29 & \textbf{92.04$\pm$0.44} \\
\textbf{FeRG-LLM}      
& \textbf{81.55$\pm$8.64} & 60.38$\pm$0.78 & 87.42$\pm$0.27 
& \textbf{83.83$\pm$0.51} & \textbf{85.76$\pm$0.16} & \textbf{90.17$\pm$0.33} & 91.75$\pm$0.30 \\
\bottomrule
\end{tabular}
\caption{The overall mean AUC and standard deviation for major models, including FeRG-LLM, are presented. XGB, the baseline model \citep{Chen:2016:XST:2939672.2939785}, achieved an AUC of 80.78 without feature engineering. After each model generated features for 14 binary classification tasks, AUC scores were compared : CAAFE scored 81.22, FeatLLM 71.51, Llama 3.1 70B 83.9, and FeRG-LLM achieved the highest at 84.79. For FeatLLM, MICE \citep{van2011mice} was applied to handle missing values, as recommended in the original paper.} 
\label{main-results}
\end{table*}

\section{Experiments}
\subsection{Experimental Setup}
To evaluate how effectively FeRG-LLM performs feature engineering, we design a method that mirrors the process data scientists follow when discovering new features. Since they refine feature through trial and error, our approach replicates their workflow, enabling us to assess the feature generation capability of LLM. The hardware setup is introduced separately in Section 4.5, along with an analysis of inference time.

\paragraph{AUC evaluation Method} 
We use various tabular datasets, each with its own ML objective, to evaluate the quality of our model-generated features. For each evaluation dataset, the experiment is repeated 7 times, measuring the mean and standard deviation for each experiment  as detailed in the pseudocode provided in Algorithm \ref{alg:eval}. For binary classification tasks, we employ XGBoost for model fitting using AUC as the evaluation metric based on the ROC curve. When fitting the ML model, we split the data into training and testing sets at an 8:2 ratio and perform simple hyperparameter tuning for XGBoost.

\begin{algorithm}[!h]
\small
\caption{AUC Evaluation Method}
\begin{algorithmic}[1]
\STATE Let $N \gets$ \COMMENT{\# of experiments}
\STATE Let $L \gets$ \COMMENT{Max iterations without improvement}
\STATE Let $K \gets$ \COMMENT{Required improvements}
\STATE Calculate $A_{\rm{basic}}$ using XGBoost on the dataset \COMMENT{Baseline AUC}
\FOR{$i \gets 1$ to $N$}
    \STATE Initialize $c \gets 0$, $t \gets 0$
    \STATE Initialize set $S_i \gets \emptyset$ \COMMENT{Set to store improved AUCs}
    \WHILE{$c < K$ \AND $t < L$}
        \STATE Obtain $A_{\rm{curr}}$ from model
        \IF{$A_{\rm{curr}} > A_{\rm{basic}}$}
            \STATE Add $A_{\rm{curr}}$ to $S_i$
            \STATE $c \gets c + 1$
            \STATE $t \gets 0$
        \ELSE
            \STATE $t \gets t + 1$
        \ENDIF
    \ENDWHILE
    \STATE Save $S_i$ for experiment $i$
\ENDFOR
\STATE Collect the maximum AUC values from \( S_1, S_2, \dots, S_N \) (i.e., \( A_{\rm{max}, i} = \max(S_i) \)) into a set \( S \), then compute its mean and standard deviation.
\end{algorithmic}
\label{alg:eval}
\end{algorithm}

\paragraph{Experiment Execution} We utilize FeRG-LLM with a prompt containing dataset and ML task information, setting the temperature to 1.4 and top-p parameter to 0.9. The model generated Python code for feature creation, which was subsequently executed to produce a dataset with the newly added features.\newline
If at most five features are generated we perform a exhaustive search to evaluate AUCs for all combinations and select the one with highest. For six or more, we evaluate each one individually as well as the entire set for efficiency. This procedure is repeated until $K$ \footnote{We conduct experiments setting $N = 7 $, $L = 15 $, and $ K = 3 $.} improved AUC outputs are observed and the best value is recorded. The model terminates after $L$ iterations if there is no improvement. 
\paragraph{Datasets}
We prepare datasets from the UCI Repository, representing standard machine learning problems, including Bank \citet{bank_marketing_222}, Nhanes \citet{national_health_and_nutrition_health_survey_2013-2014_(nhanes)_age_prediction_subset_887}, Cultivar \citet{forty_soybean_cultivars_from_subsequent_harvests_913}, Ilpd-indian \citet{ilpd_(indian_liver_patient_dataset)_225}, Glioma \cite{glioma_grading_clinical_and_mutation_features_759}, Support2 \citet{knaus1995support}, and Heart-failure \citet{heart_failure_clinical_records_519}. Additionally, we obtain datasets from Kaggle, including Diabetes, Credit, Ethereum, Churn, and Ad-click, to assess their applicability in real-world classification tasks. Also, there is a potential risk that the dialouge dataset might contain direct information about the evaluation sets. To address this, we additionally include the recent datasets : Aging-npha \cite{national_poll_on_healthy_aging_(npha)_936} and  Infrared-t \cite{Wang2021} from the UCI Repository. Detailed information on each dataset, including data size, label ratio, and the specific machine learning problem, is provided in Appendix \ref{sec:appendixE}.
\subsection{Main Results}

\begin{table*}[!h]
\centering
\small
\renewcommand{\arraystretch}{1.5} 
\begin{tabular}{p{2.3cm}|p{13cm}}
\Xhline{1pt}
\textbf{ML task\newline Prompt \newline(NHANES)} & [Domain] Medical\quad [Type] classification\newline

[Present variable in dataframe]\newline 
riagendr: Respondents Gender,\newline
paq605:If the respondent engages in moderate or vigorous-intensity sports,\newline fitness, or recreational activities in the typical week,\newline
bmxbmi: Respondents Body Mass Index,\newline
lbxglu: Respondents Blood Glucose after fasting,\newline
diq010:If the Respondent is diabetic,\newline
lbxglt: Respondents Oral ,\newline
lbxin: Respondents Blood Insulin Levels\newline

Provide key ideas for creating new derived variables to improve the performance of the following ML problem using the given dataset.\newline

[Machine Learning Problem] for predicting the given person’s age group from the record. \\
\hline
\hline
\textbf{Without DPO \newline alignment  \newline Rationale } & definition: interaction term between engagement in activities and bmi, useful to assess the joint effect of these two features on aging risk.\newline
consider relationships between physical activity (as indicated by paq605), body metrics (such as bmxbmi), and glucose levels (lbxglu) could help model aging risks as they interact to influence age-related factors.\newline\newline
definition: average glucose levels adjusted for diabetic status, helping to evaluate the diabetic effects on glucose.\newline
interaction features can provide insights into how these different metrics relate to aging, enabling improved predictions. \\
\hline
\textbf{Python code} & df['activity\_bmi\_interaction'] = df['paq605'] * df['bmxbmi']\newline
df['adjusted\_avg\_glucose'] = df['lbxglu'].where(df['diq010'] == 1, 0) \\
\hline
\textbf{AUC } & \textbf{75.31} \\
\hline
\hline
\textbf{FeRG-LLM \newline Rationale} & definition: the ratio of body mass index (bmi) to blood glucose levels, which might indicate weight management relative to glucose status.\newline
the features represent a combination of health metrics that can be interrelated. for example, body mass index (bmi) in relation to blood glucose and activity levels might shed light on overall health status more effectively than examining each variable in isolation.\newline

definition: the average of bmi and blood glucose levels to understand overall health status more effectively.\newline
interaction between bmi and blood glucose might provide insights into the relationship between weight and glucose metabolism.\newline

definition: the ratio of activity levels to blood glucose, depicting how physical activity might relate to glucose status.\newline
ratios or combinations of certain health metrics may better capture the nuances of an individual's health profile, which could improve prediction accuracy for age groups. \\
\hline
\textbf{Python Code} &

df['bmi\_glucose\_ratio'] = df['bmxbmi']/df['lbxglu']\newline
\footnotesize{df['average\_health\_metric'] = (df['bmxbmi'] + df['lbxglu'] + df['diq010'].replace({'yes':1, 'no':0}))/3}\newline
df['activity\_glucose\_ratio'] = df['paq605']/df['lbxglu']
\\
\hline
\textbf{AUC} & \textbf{76.75} \\
\Xhline{1pt}
\end{tabular}
\caption{Linguistic Analysis of Reasoning Shift Induced by Rationale Alignment. In the NHANES task, examining the features and their corresponding evidence generated using variables such as paq605, bmxbmi, lbxglu, and diq010 reveals significant changes before and after DPO application. }
\label{tab:scene_table}
\end{table*}

\paragraph{Baselines} 
We evaluate our method with CAAFE \cite{hollmann2024large}, FeatLLM \cite{han2024large} and Llama 3.1 70B. In the original framework of CAAFE, TabPFN \cite{hollmann2025tabpfn} is used for model training and providing criteria such as AUC and ACC for feature selection. However, we replace it with XGBoost because TabPFN struggled to handle large datasets such as Bank and Ethereum, and often fell short compared to XGBoost in various experiments. FeatLLM generates features with a GPT-based API using 16 data points and trains a linear layer on up to 128 data points. To further investigate the model's capabilities, we test a setup where FeatLLM generated features using 64 data points instead of 16, and the linear layer is trained on the entire dataset instead of 128 data points. After evaluating multiple configurations, we report the highest observed AUC.\newline
Furthermore, we include the Llama 3.1 70B model, which has approximately nine times more parameters than the 8B model, offering significantly greater capacity for reasoning and knowledge retention. This inclusion allows for a meaningful comparison between a large-scale model and a more resource-efficient fine-tuned alternative. All results are presented in Appendix \ref{appendix:caafe_ferg}.

\begin{figure}[H]
    \centering
    \includegraphics[width=1\linewidth]{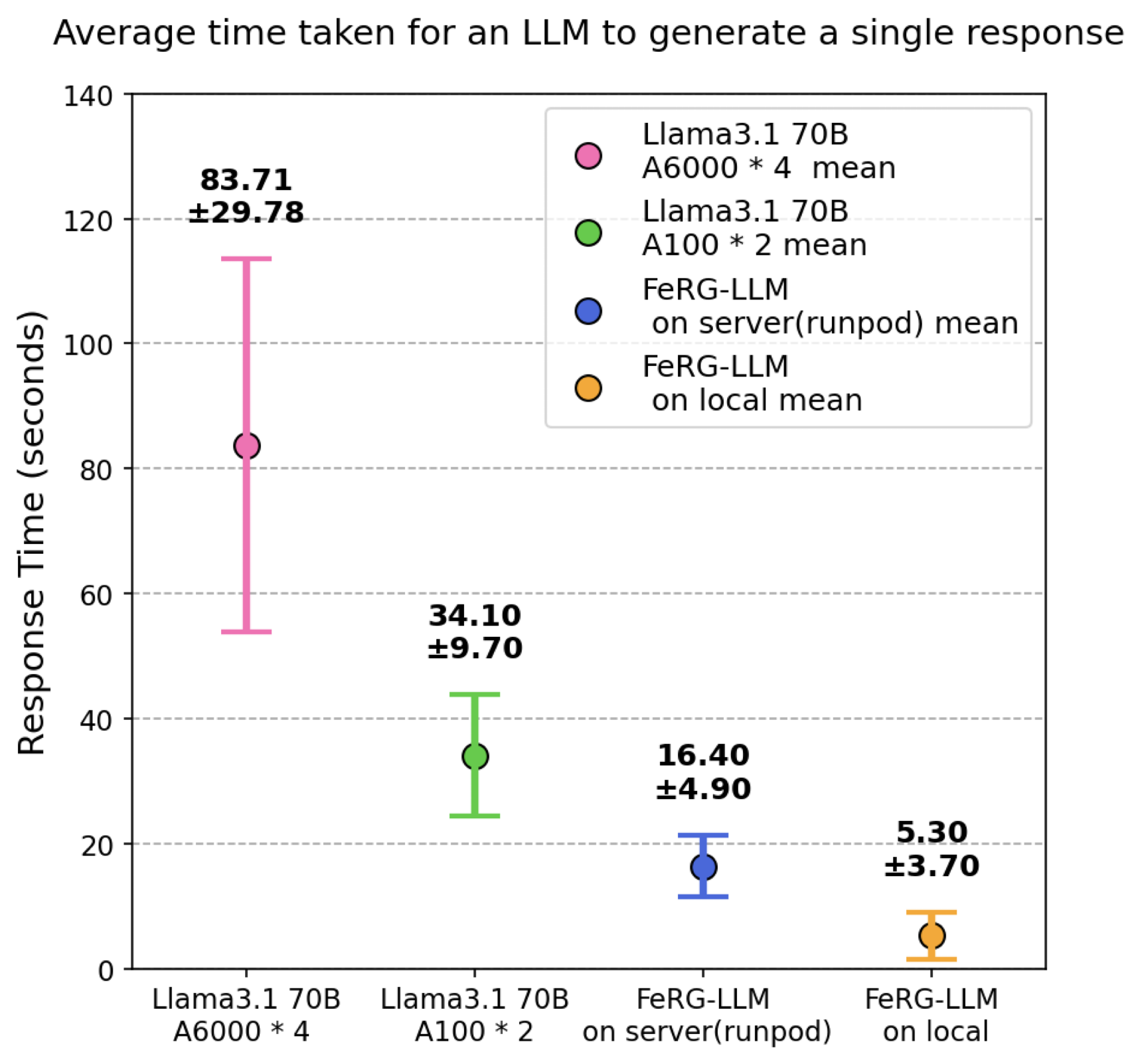}
    \caption{Inference time comparison of the 70B model and FeRG-LLM using different configurations. The 70B model takes 83.71 seconds on four RTX A6000 and 34.10 seconds on two A100, while FeRG-LLM completes inference in 16.4 seconds on a single RTX A6000 and local feature generation in about 5 seconds on a single RTX A6000.}
    \label{fig:time}
\end{figure}

\paragraph{Performance Comparison} 
Table \ref{main-results} presents overall AUC evaluation results. FeRG-LLM achieves the highest performance on 7 datasets and the second-highest on the remaining 7, surpassing other recent models. Notably, it performs on par with the 70B model and even outperforms it in Ad-click and Cultivar task. This implies that FeRG framework can deliver performance comparable to that of larger language models while operating with fewer resource constraints and at greater speed, as shown in Figure \ref{fig:time}. Meanwhile, relatively low AUC have at times been identified in FeatLLM, suggesting that a single linear layer may be insufficient to capture the underlying structure of the data. Conclusively, unlike prior research models that rely on external APIs to generate features, our model functions locally, ensuring data security in corporate environments.

\subsection{Reasoning Shift Analysis}
We study the effect of providing feedback on generated reasoning by visualizing representations of rationales through t-SNE, leading to linguistic analysis of reasoning shift. 

\paragraph{Linguistic Analysis of Response} We conduct a linguistic analysis of the feature generation rationale and the final code as shown in Table \ref{tab:scene_table}. For NHANES task, FeRG-LLM and the model prior to DPO both adopt the same four variables paq605, bmxbmi, lbxglu, and diq010, yet yield entirely distinct outputs. This shift improve AUC from 75.31 to 76.75. We also confirm that the Python code is derived directly from the rationale, indicating a strong link between the explanation and final implementation. More examples of FeRG-LLM's responses are in Appendix \ref{sec:appendix_examples}.
\begin{figure}[!h]
    \centering
    \includegraphics[width=1\linewidth]{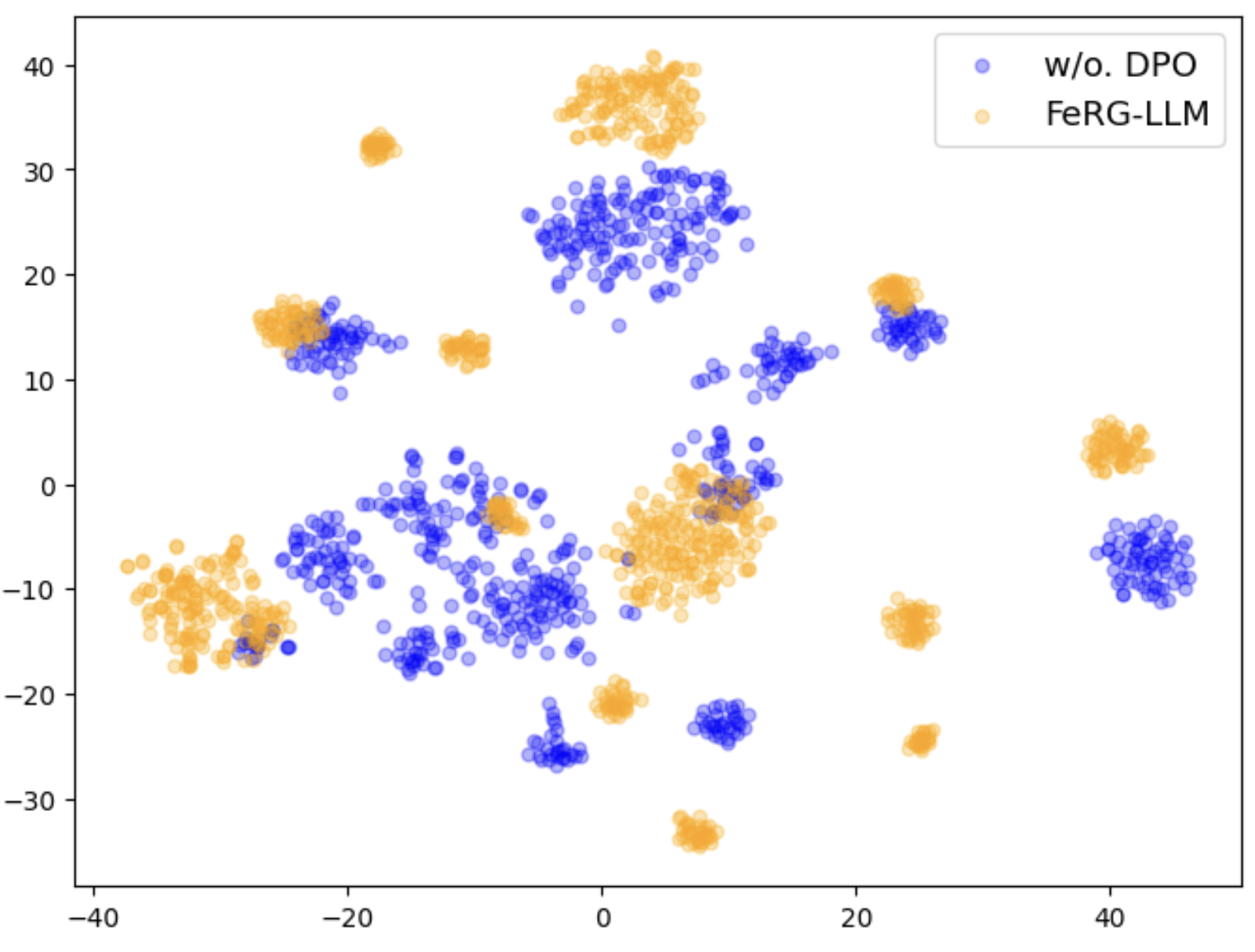}
    \caption{t-SNE Visualization of Reasoning Embeddings. Yellow points represent embeddings of reasoning generated by FeRG-LLM, while blue points correspond to those generated without DPO.}
    \label{fig:tSNE}
\end{figure}

\paragraph{Embedding Shift} To explore the vector representations of each response, we use OpenAI text-embedding-3-small model. As shown in Figure \ref{fig:tSNE}, the reason distribution without alignment (blue) and that of FeRG-LLM (orange) exhibit distinct shifts in reasoning patterns. After applying DPO feedback, new content emerged where none had existed before, while previously produced content converged into a more narrowly confined space.

\paragraph{Ablation Study} Table \ref{tab:ablation} presents the ablation results for FeRG-LLM across 14 datasets. In every case, generating features based on rationales contributes to superior performance compared to their absence. On average, the model that immediately produces Python code exhibits a 1.06\% lower AUC than its rationale-driven counterpart.\newline
In addition, the effectiveness of alignment is quantitatively confirmed with a 0.34\% AUC gain. While this improvement may seem modest, the consistently better performance across most datasets suggests that DPO meaningfully enhances the expressiveness of the generated features.

\begin{table}[!h]
\centering
\small
\renewcommand{\arraystretch}{1.5} 
\resizebox{0.98\linewidth}{!}{
\begin{tabular}{c|ccc}
\Xhline{1pt}
Dataset & w/o. Rationale & w/o. DPO & FeRG-LLM \\
\hline
Bank & 93.69$\pm$0.44 & 93.40$\pm$0.10 & \textbf{93.88$\pm$0.51} \\
\hline
Nhanes & 75.84$\pm$0.39 & 75.73$\pm$0.34 & \textbf{76.23$\pm$0.40} \\
\hline
Diabetes & 82.25$\pm$0.51 & 82.25$\pm$0.29 & \textbf{82.43$\pm$0.34} \\
\hline
Cultivar & 76.57$\pm$2.36 & \textbf{79.94$\pm$5.24} & 78.34$\pm$2.74 \\
\hline
Credit & 83.93$\pm$0.59 & 84.11$\pm$0.38 & \textbf{85.07$\pm$0.67} \\
\hline
Ethereum & 96.76$\pm$0.50 & 96.78$\pm$0.92 & \textbf{97.03$\pm$0.60} \\
\hline
Churn & 93.04$\pm$0.45 & 93.03$\pm$0.30 & \textbf{93.22$\pm$0.14} \\
\hline
Ad-click & 76.22$\pm$9.72 & 80.50$\pm$12.65 & \textbf{81.55$\pm$8.64} \\
\hline
Aging-npha & 59.85$\pm$1.44 & 60.36$\pm$1.13 & \textbf{60.38$\pm$0.78} \\
\hline
Infrared-t & 87.12$\pm$0.28 & 87.20$\pm$0.23 & \textbf{87.42$\pm$0.27} \\
\hline
Ilpd-indian & 82.26$\pm$1.02 & 82.73$\pm$0.76 & \textbf{83.83$\pm$0.51} \\
\hline
support2 & 85.66$\pm$0.10 & 85.65$\pm$0.11 & \textbf{85.76$\pm$0.16} \\
\hline
heart-failure & 89.63$\pm$0.40 & 89.79$\pm$0.19 & \textbf{90.17$\pm$0.33} \\
\hline
glioma & 91.58$\pm$0.12 & 91.53$\pm$0.07 & \textbf{91.75$\pm$0.30} \\
\hline
\textbf{Mean} & \makecell{83.89 \\ (-1.06\%)} & \makecell{84.50 \\ (-0.34\%)} & \makecell{\textbf{84.79} \\ \textbf{(--)}} \\

\Xhline{1pt}
\end{tabular}
}
\caption{Ablations of each component of FeRG-LLM : Rationle prompting and dpo alignment process.}
\label{tab:ablation}
\end{table}

\subsection{Beyond Classification Task}
\begin{table}[!h]
\centering
\scriptsize
\renewcommand{\arraystretch}{1.5} 
\begin{tabular}{>{\centering\arraybackslash}p{1.1cm}|>{\centering\arraybackslash}p{1.3cm}>{\centering\arraybackslash}p{1.8cm}>{\centering\arraybackslash}p{1.7cm}}
\Xhline{1pt}
Dataset & XGBReg & w/o. DPO & \textbf{FeRG-LLM} \\
\hline
\textbf{Bike}       & 1408.80 & 1368.63 $\pm$ 9.50 \newline (-2.85\%) & 1337.76 $\pm$ 35.35 \newline (-5.04\%) \\
\textbf{Concrete}   & 18.61   & 17.18 $\pm$ 0.66 \newline (-7.68\%) & 16.47 $\pm$ 0.58 \newline (-11.50\%) \\
\textbf{Parkinsons} & 3.04    & 2.69 $\pm$ 0.09 \newline (-11.51\%) & 2.76 $\pm$ 0.05 \newline (-9.21\%) \\
\textbf{Realestate} & 32.60   & 30.18 $\pm$ 0.94 \newline (-7.42\%) & 29.26 $\pm$ 0.81 \newline (-10.24\%) \\
\Xhline{1pt}
\end{tabular}
\caption{Overall Mean MSE and Standard Deviation for Regression Tasks Using FeRG-LLM.}
\label{tab:reg-results}
\end{table}

We further highlight that our framework can be extended to other tasks. Specifically we apply FeRG-LLM to regression datasets, details in Appendix \ref{sec:regdata}, by switching the evaluation metric from AUC to MSE and replacing XGBClassifier with XGBRegressor. Table \ref{tab:reg-results} presents a 5–10\% decrease in MSE across all datasets compared to cases without feature engineering. Also, FeRG-LLM generally outperforms a non-rationale baseline on regression tasks—except for the Concrete—demonstrating the efficacy of DPO-driven feedback. Hence, our approach is not limited to classification tasks but has the potential to improve performance over a wider spectrum of predictive problems.

\subsection{Inference Speed}

In Section 4.2, FeRG-LLM and Llama 3.1 70B alternate in achieving top performance. Therefore, to further assess FeRG-LLM’s efficiency, we measure inference speed by generating 20 outputs for each of the 12 datasets, resulting in a total of 240 runs. This approach ensures that the results are relatively consistent across various datasets. The Llama 3.1 70B model run on an AMD EPYC 7513 system with 503 GB of RAM in two configurations: four RTX A6000 GPUs (48GB VRAM each) and two A100 GPUs (80GB VRAM each). FeRG-LLM is tested on a university-shared Intel Xeon Gold 6348 server with 16GB allocated per user and a single RTX A6000. It is also tested on RunPod with the same system specifications as the 70B model, using one RTX A6000 with 48 GB of VRAM. 
On the cloud, the 70B model takes an average of 83.71 seconds in the A6000 setup and 34.10 seconds in the A100 setup. But FeRG-LLM is much faster, averaging 16.4 seconds and only 5.3 seconds in non-cloud settings. In summary, given the similar performance of both models and the results from experiments conducted according to Algorithm \ref{alg:eval}, where FeRG-LLM completed an average of 7.4 runs per experiment compared to 5.7 for the 70B model, our model achieves significantly higher efficiency with fewer GPUs and faster inference time.

\section{Conclusion}

In this research, we develop a novel framework, FeRG-LLM, an automated feature engineering language model that assists data scientists in discovering features for building effective machine learning models. By optimizing a small LLM (e.g., Llama 3.1 8B), FeRG-LLM can operate efficiently even on a company's limited infrastructure and ensure data security. This study explores LLM capabilities by pushing their limits to assess their potential in automating data-driven tasks.\newline
The proposed framework offers significant practical utility in real-world industrial applications. It is ideal for startups and teams with limited GPU resources and workforce. Even large enterprises can achieve automation and efficiency by applying our model to large-scale datasets with a large number of variables. Also, our model can be easily adapted to diverse tasks by simply replacing the ML prompts with appropriate tasks and evaluation metric. This flexibility makes the framework highly extensible and opens avenues for future research in diverse machine learning contexts.

\section*{Limitation}

A limitation of our study is the insufficient analysis of the trade-off between computational overhead and performance gains in our two-step chain-of-thought and DPO training process. Although DPO-driven feedback improves performance compared to the model before DPO training, the additional training and inference costs are not fully quantified. Future work should assess these trade-offs to better establish the practical viability of our approach.\newline
From a structural perspective, our model performs two sequential calls to leverage the LLM's Chain-of-Thought capability. However, it currently generates features directly from the initial reasoning output without further refinement. Additionally, our investigation into the optimal step size remains insufficient. These limitations warrant further exploration and improvement in future research.

\newpage
\appendix
\onecolumn
\section{Details : Data Generation from papers}
\label{sec:appendixA}
We utilize OpenAI GPT-4o-mini to gather essential information from various papers related to tabular data analysis. Figure \ref{fig:data-generate-prompt} illustrates the specific prompt used for data extraction within these studies.\\
A total of 7 sections are extracted titled `Domain’, ‘Type’, ‘Available Features in dataframe’, ‘ML Problem’, ‘Thought for deriving new feature’, ‘New Features’, and ‘Python code.’, and the final Python code is provided without any additional explanations or text, allowing it to be executed directly. Subsequently, we make numerous revisions to enhance the quality of the data, ensuring its accuracy and reliability.

\begin{figure*}[!h]
\centering
\begin{tcolorbox}[width=\linewidth, colframe=gray!50, colback=white, boxrule=1.2pt, arc=7pt]
\spaceskip=0.4em
\small \ttfamily

{\small\textbf{System Instruction:}}

\tikz \draw[dash pattern=on 3pt off 1pt] (0,0) -- (0.5\linewidth,0);

\vspace{10pt}
You are a highly skilled machine learning engineer. You need to extract the content used to create new derived features to improve actual machine learning performance based on the information in the paper.

\vspace{9pt}
1. Read the provided text and organize it into the following sections:\textcolor{dimgray}{`Domain', `Type',} \textcolor{dimgray}{`Available Features in dataframe', `ML Problem', `Thought for deriving new feature', `New} \textcolor{dimgray}{Features'}, and \textcolor{dimgray}{`Python code.'}

\vspace{7pt}
2. Since the focus is on discovering new features applicable to existing tabular data, if there is no appropriate information for each section, refer to previous prompts or simply answer "None" for that section.

\vspace{7pt}
3. Include the variable names and their descriptions from the tabular dataset in the \textcolor{dimgray}{`Available Features in dataframe'} section.

\vspace{7pt}
4. If new derived features are mathematically defined, ensure the variables in the formula come from the \textcolor{dimgray}{`Available Features in dataframe'} section only.

\vspace{7pt}
5. Even if there is no information on new derived features in the text, if you can propose features based solely on the \textcolor{dimgray}{`ML Problem'} and \textcolor{dimgray}{`Available Features in dataframe,'} you may do so in the \textcolor{dimgray}{'New Features'} section, even if it's not from the paper.

\vspace{7pt}
6. Include the ideas, knowledge, and expertise used to create new derived features to improve machine learning performance in the \textcolor{dimgray}{`Thought for deriving new feature'} section.

\vspace{7pt}
7. Assume the contents in \textcolor{dimgray}{`Available Features in dataframe'} are columns present in a DataFrame `df,' and convert the formulas in \textcolor{dimgray}{`New Features'} into Python code. Respond with only the Python code. It should be executable.

\vspace{7pt}
8. If there are multiple contexts or solutions, record them separately.

\vspace{9pt}
Please respond in English.

\vspace{10pt}
\end{tcolorbox}
\caption{Data Generation Prompt Using the GPT API.}
\label{fig:data-generate-prompt}
\end{figure*}

\clearpage

\section{More Example of Feature Engineering with FeRG-LLM}
\label{sec:appendix_examples}
The results of FeRG-LLM's feature engineering for Ethereum and the Cultivar task are presented in the table below. As discussed in the main text the model identifies a rationale to guide the discovery of new features and subsequently generates the final Python code based on this rationale.
\begin{table*}[htbp]
\renewcommand{\arraystretch}{1.5} 
\centering
\scriptsize
\begin{tabular}{p{2.0cm}|p{13.3cm}}

\toprule
\scriptsize

\textbf{System\newline Instruction} & You are an automated ML engineer who writes Python code based on a given domain, present variables in a dataframe, and ML problems. Whenever the sentence write down python code appears, immediately write Python code. \\
\midrule
\textbf{ML task\newline Prompt \newline(Ethereum)} & [Domain] Blockchain\quad [Type] classification\newline

[Present variable in dataframe]\newline 
txhash: Unique Transaction Hash

blockheight: The block number in which the transaction was included

timestamp: The exact date and time when the transaction was processed

from: The sender's Ethereum address

to: The receiver's Ethereum address

value: The amount of Ether transferred in the transaction

Provide key ideas for creating new derived variables to improve the performance of the following ML problem using the given dataset.\newline

[Machine Learning Problem] identify patterns and anomalies indicative of fraudulent activity. \\
\midrule
\textbf{FeRG-LLM \newline Rationale } & 
1. definition: total amount transferred by a sender within a given time frame.\newline
- aggregated values to assess sender-receiver behavior, which can help identify unusual patterns associated with fraud.\newline
2. definition: count of distinct transactions initiated by a particular sender in a specific time frame.\newline
- use of time-based features to uncover patterns over time and evaluate frequency and volume of transactions, allowing for better fraud detection.\newline
3. definition: total number of times a unique sender-receiver pair exchanged values across transactions\newline
- utilize transaction volume and frequency data to identify abnormal flows indicative of potential scams. \\
\midrule
\textbf{Python code} & \# 1. total transaction value sent by each sender over a certain period.\newline
df['totalvalue'] = df.groupby('from')['value'].transform('sum')\newline
\# 2. count of transactions per sender within a specific time frame.\newline
df['count\_transactions'] = df.groupby('from')['from'].transform('count')\newline
\# 3. frequency of transaction for sender-receiver pairs.\newline
df['frequency\_tx'] = df.groupby(['from','to'])['from'].transform('count') \\
\hline
\hline
\textbf{ML task\newline Prompt \newline(Cultivar)}&
[Domain] Agriculture\quad [Type] classification\newline

[Present variable in dataframe]\newline 
season: 1 or 2

cultivar: Cultivar names

repetition: 1, 2, 3 or 4

ph: plant height (cm), determined from the soil surface to the insertion of the last leaf using a millimeter ruler

ifp: insertion of the first pod (cm), determined from the soil surface to the insertion of the first vegetable

nlp: Number of stems (unit), through manual counting

ngp: Number of legumes per plant (unit), through manual counting

ngl: Number of grains per plant (unit), through manual counting

ns: Number of grains per pod (unit), through manual counting

mhg: Thousand seed weight (g), according to the methodology described in Brasil (2009)

Provide key ideas for creating new derived variables to improve the performance of the following ML problem using the given dataset.\newline

[Machine Learning Problem] for predicting how high the grain yield of this soybean cultivar will be. \\
\midrule
\textbf{FeRG-LLM \newline Rationale } & 
1. definition: average number of pods per plant considering the possible repetition of experiments for different seasons and cultivar replications.\newline
   • purpose of average biomass measures: the various measurements related to plant size and development can be used to compute derived metrics which might relate more closely to grain yield.\newline
2. definition: variation in the number of grains produced per plant across different experimental replications which can highlight stability and performance.\newline
   • purpose of pod-to-grain ratios: examining the relationship between the number of pods and grains could provide insights into the efficiency of pod utilization, potentially revealing cultivar performance characteristics.
 \\
\midrule
\textbf{Python code} & 
df['avgpp'] = (df['ngp'].where(df['repetition'] == 1).fillna(0) + df['ngp'].where(df['repetition'] == 2).fillna(0) +\newline
df['ngp'].where(df['repetition'] == 3).fillna(0) + df['ngp'].where(df['repetition'] == 4).fillna(0)) / 4\newline
df['grv'] = df.groupby('cultivar')['ngl'].transform(lambda x: ((x - x.mean())**2).mean())
\\
\bottomrule
\end{tabular}
\caption{\small{Example responses when Ethereum and Cultivar task are given into FeRG-LLM. Definitions are generated for each inference, followed by the corresponding Python code for new features. Some responses that generate code include comments indicated by "\#" while others do not. In both cases the Python code can be executed immediately, ensuring that automation works properly.}}
\label{tab:appendix_examples_table}
\end{table*}

\clearpage
\section{All Experimental Results for CAAFE and FeatLLM}
\label{appendix:caafe_ferg}
In CAAFE, variable selection is performed using TabPFN instead of XGBoost, and the final dataset is fitted with XGBoost. As discussed in the main text, the model fails to fit properly on the Bank and Ethereum datasets, which contain 45,000 and 254,973 samples, respectively. As a result, AUC and ACC values could not be computed.
\begin{table*}[h!]
\small
\renewcommand{\arraystretch}{1.3}
\centering
\begin{tabular}{l|c}
\Xhline{1pt}
Dataset       & CAAFE+TabPFN \\ \hline\hline
Bank         & NaN                              \\ \hline
Nhanes        & $75.01 \pm 0.68$                 \\ \hline
Diabetes      & $80.47 \pm 1.11$                 \\ \hline
Cultivar      & $71.43 \pm 4.93$                 \\ \hline
Credit        & $82.01 \pm 0.91$                 \\ \hline
Ethereum      & NaN                              \\ \hline
Churn         & $92.38 \pm 0.39$                 \\ \hline
Ad-click      & $63.85 \pm 0.00$                 \\ \hline
Aging-npha    & $56.40 \pm 0.81$                 \\ \hline
Infrared-t    & $86.04 \pm 0.35$                 \\ \hline
Ilpd-indian   & $80.53 \pm 0.98$                 \\ \hline
Support2      & $85.38 \pm 0.05$                 \\ \hline
Heart-failure & $87.63 \pm 2.23$                 \\ \hline
Glioma        & $90.79 \pm 0.30$                 \\
\Xhline{1pt}
\end{tabular}
\caption{Performance results of the CAAFE+tabpfn model on various datasets.}
\label{tab:performance}
\end{table*}

When testing FeatLLM, we use a setup where FeatLLM generated features using 16 data points(shots) or 64, and the linear layer is trained on 128 data points or the entire dataset. 

\begin{table*}[!h]
\centering
\small
\renewcommand{\arraystretch}{1.2} 
\begin{tabular}{c|cccc}
\Xhline{1pt}
Dataset & All data (16 shots) & All data (64 shots) & 128 data points (16 shots) & 128 data points (64 shots) \\
\hline
\hline
Bank & 69.25$\pm$1.17 & 66.62$\pm$1.23 & 80.05$\pm$0.62 & 79.17$\pm$0.75 \\
Nhanes & 59.47$\pm$3.89 & 60.95$\pm$1.23 & 65.73$\pm$0.49 & 66.57$\pm$0.71 \\
Diabetes & 78.66$\pm$0.47 & 78.16$\pm$0.24 & 78.45$\pm$0.36 & 78.53$\pm$0.38 \\
Cultivar & 63.19$\pm$2.63 & 62.21$\pm$2.93 & 69.52$\pm$2.23 & 63.12$\pm$3.00 \\
Credit & 50.64$\pm$7.19 & 36.30$\pm$1.45 & 36.09$\pm$1.12 & 40.21$\pm$2.55 \\
Ethereum & 73.19$\pm$0.98 & 75.48$\pm$2.05 & 91.22$\pm$0.73 & 87.61$\pm$2.65 \\
Aging\_npha & 58.52$\pm$1.22 & 59.16$\pm$0.75 & 57.47$\pm$1.34 & 56.03$\pm$0.76 \\
Infrared\_t & 80.48$\pm$0.53 & 79.11$\pm$0.81 & 79.04$\pm$0.81 & 77.03$\pm$0.98 \\
Churn & 73.71$\pm$3.28 & 77.48$\pm$2.38 & 79.91$\pm$0.79 & 79.38$\pm$1.16 \\
Ad\_click & 51.18$\pm$0.57 & 48.98$\pm$0.46 & 47.72$\pm$0.39 & 47.30$\pm$0.59 \\
Ilpd\_indian & 82.61$\pm$1.17 & 81.74$\pm$0.81 & 80.67$\pm$0.71 & 81.13$\pm$0.69 \\
Support2 & 56.70$\pm$7.96 & 55.74$\pm$7.99 & 47.31$\pm$7.68 & 47.42$\pm$4.53 \\
Heart\_failure & 78.26$\pm$0.91 & 80.77$\pm$1.08 & 77.55$\pm$1.28 & 78.36$\pm$1.96 \\
Glioma & 70.13$\pm$5.57 & 70.59$\pm$5.98 & 73.51$\pm$3.28 & 65.59$\pm$6.87 \\
\Xhline{1pt}
\end{tabular}
\caption{Performance results of the FeatLLM on the various setups.}
\label{tab:performance}
\end{table*}

\clearpage

\section{Training Setup and XGBoost Parameter Search}
\label{sec:trainsetup}

\subsection{Training Setup}
We detail our training configurations for both the supervised fine-tuning (SFT) and Direct Preference Optimization (DPO) settings.
For SFT, we use a batch size of 4 and train the model for 5 epochs with a learning rate of $1\times10^{-4}$. Mixed precision training is performed using BF16, and LoRA is applied with a rank of 8, an alpha of 16, and no dropout (0.0). For both SFT and DPO, we utilize a cosine learning rate scheduler. In the case of DPO, the model is trained with a learning rate of $8\times10^{-6}$ for 4 epochs, and the $\beta$ parameter is set to 0.1.

\begin{table}[h]
\centering
\begin{tabular}{lcc}
\toprule
\textbf{Parameter} & \textbf{SFT} & \textbf{DPO} \\ \midrule
Batch Size         & 4            & 4          \\
Epochs             & 5            & 4          \\
Learning Rate      & $1\times10^{-4}$ & $8\times10^{-6}$ \\
Mixed Precision    & BF16         & BF16       \\
LoRA Rank          & 8            & 8          \\
LoRA Alpha         & 16           & 16         \\
LoRA Dropout       & 0.0          & 0.0        \\
Scheduler          & Cosine       & Cosine     \\
$\beta$               & --           & 0.1        \\ \bottomrule
\end{tabular}
\caption{Training configurations for SFT and DPO with a cosine learning rate scheduler.}
\label{tab:training_setup}
\end{table}
\vspace{1in}
\subsection{XGBoost Parameter Search}
For fitting the XGBoost model, we conducted a grid search over the following parameter ranges:

\begin{table}[h]
\centering
\begin{tabular}{ll}
\toprule
\textbf{Parameter}       & \textbf{Values}   \\
\midrule
max\_depth               & 3, 5, 7           \\
learning\_rate           & 0.01, 0.1         \\
n\_estimators            & 50, 100, 200      \\
subsample                & 0.8, 1.0          \\
colsample\_bytree        & 0.8, 1.0          \\
\bottomrule
\end{tabular}
\caption{Hyperparameter search space for XGBoost.}
\label{tab:xgb_params}
\end{table}

\clearpage

\section{Datasets for Feedback}
\label{sec:appendixD}

These datasets contain information on target datasets for which features were generated using the supervised fine-tuned model. Based on the evaluation results, we utilized the final reject and chosen datasets to create the training data used for DPO.
\begin{table*}[!h]
\renewcommand{\arraystretch}{1.5} 
\small
\centering
\begin{tabular}{>{\centering\arraybackslash}m{2.3cm}cccm{6.5cm}}
\Xhline{1pt}
\textbf{Dataset} & \textbf{Size} & \textbf{\# Features} &\textbf{Label Ratio(\%)} & \textbf{ML Problem}\\
\hline
\hline
Default of credit card clients & 30000 & 23 & 77.9 $\backslash$ 22.1 & Predicting customer default payments in Taiwan by comparing the predictive accuracy of the probability of default across six data mining methods.\\
\hline
In vehicle coupon recommendation & 12684 & 23 & 56.8 $\backslash$ 43.2 & Predicting whether an individual will accept a recommended coupon under different driving scenarios.
\\
\hline
Online news popularity & 39644 & 58 & 85.2 $\backslash$ 14.8 & Predicting the social media popularity of Mashable articles using a heterogeneous set of features collected over a two-year period.\\
\hline
Online shoppers purchasing intention dataset & 12330 & 17 & 84.5 $\backslash$ 15.5 & Predicting whether a session led to a purchase.\\
\hline
Polish companies bankruptcy & 43405 & 65 & 95.2 $\backslash$ 4.8 & Predicting bankruptcy of Polish companies based on financial data, where bankrupt companies were analyzed from 2000 to 2012 and operating companies were evaluated from 2007 to 2013.\\
\hline
Blood transfusion service center & 748 & 4 & 76.2 $\backslash$ 23.8 & Predicting whether a blood donor will donate blood in March 2007 based on the donor's historical data.\\
\hline
Iranian churn & 3150 & 13 & 84.3 $\backslash$ 15.7 & Predicting customer churn in the telecom industry based on user activity and subscription patterns over a 12-month period.\\
\hline
Productivity prediction of garment employees & 1197 & 14 & 50.2 $\backslash$ 49.8 & Predicting the productivity of garment employees based on various factors.\\
\Xhline{1pt}
\end{tabular}
\caption{\label{dataset_description} This table provides descriptions of the datasets used for DPO Training. The reasons for preference based on the AUC are ordered for each dataset.}
\end{table*}
\newpage

\section{Evaluation Datasets for Classification}
\label{sec:appendixE}

\begin{table*}[!h]
\renewcommand{\arraystretch}{1.5}

\centering
\small
\begingroup
\begin{tabular}{>{\centering\arraybackslash}m{1.5cm}cccm{7.2cm}}
\Xhline{1pt}
\textbf{Dataset} & \textbf{Size} & \textbf{\# Features} &\textbf{Label Ratio(\%)} & \textbf{ML Problem}\\
\hline
\hline
Bank & 45,211 & 16 & 88.3 $\backslash$ 11.7 & Whether a customer will subscribe to a term deposit.\\
\hline
Nhanes & 2,278 & 7 & 84.0 $\backslash$ 16.0 & Predicting the given person’s age group from the record.\\
\hline
Diabetes & 768 & 8 & 65.1 $\backslash$ 34.9 & Predicting the presence of diabetes in an individual. (\small\url{https://kaggle.com/datasets/uciml/pima-indians-diabetes-database})\\
\hline
Cultivar & 320 & 10 & 50.0 $\backslash$ 50.0 & Predicting how high the grain yield of this soybean cultivar will be.\\
\hline
Credit & 1,548 & 18 & 88.7 $\backslash$ 11.3 & Predicting whether the applicant is approved or rejected. \newline(\small\url{https://www.kaggle.com/datasets/rohitudageri/credit-card-details?select=Credit_card_label.csv})\\
\hline
Ethereum & 254,973 & 6 & 93.9 $\backslash$ 6.1 & Identifing patterns and anomalies indicative of fraudulent activity.\newline (\small\url{https://www.kaggle.com/datasets/chaitya0623/ethereum-transactions-for-fraud-detection})\\
\hline
Churn & 4,250 & 20 & 85.9 $\backslash$ 14.1 & Predicting whether a customer is likely to churn based on the other features. (\small\url{https://www.kaggle.com/datasets/mustafakeser4/bigquery-churn-dataset})\\
\hline
Ad-click & 10,000 & 8 & 65.0 $\backslash$ 35.0 & Predicting whether a user will click on an online ad based on their demographics, browsing behavior, the context of the ad's display, and the time of day. (\small\url{https://www.kaggle.com/datasets/marius2303/ad-click-prediction-dataset})\\
\hline
Aging-npha & 714 & 14 & 70.4 $\backslash$ 29.6 & Predicting the categorized number of doctors a patient has visited based on their health status, sleep issues, and demographic information.\\
\hline
Infrared-t & 1,020 & 33 & 59.5 $\backslash$ 40.5 & Predicting the oral temperature using the environment information as well as the thermal image readings. \\
\hline
Ilpd-indian liver patient dataset & 583 & 10 & 71.4 $\backslash$ 28.6 & The prediction task is to determine whether a patient suffers from liver disease based on the information about several biochemical markers, including albumin and other enzymes required for metabolism.\\
\hline
Support2 & 9,105 & 42 & 68.1 $\backslash$ 31.9 & The goal is to determine these patients' 2- and 6-month survival rates based on several physiologic, demographics, and disease severity information.\\
\hline
Heart-failure & 299 & 12 & 67.9 $\backslash$ 32.1 & Predicting whether a patient will die during the follow-up period.\\
\hline
Glioma & 839 & 23 & 58.0 $\backslash$ 42.0 & Predicting whether the glioma grade is classified as \"lgg\" (lower grade glioma) or \"gbm\" (glioblastoma multiforme), where 0 represents lgg and 1 represents gbm.\\
\Xhline{1pt}
\end{tabular}
\endgroup
\caption{\label{dataset_description} This table provides descriptions of the datasets used for experiments of feature engineering. These datasets are intended for solving binary classification tasks.}
\end{table*}

\clearpage

\section{Evaluation Datasets for Regression}
\label{sec:regdata}
\begin{table*}[!h]
\small
\centering
\renewcommand{\arraystretch}{1.3}
\begingroup
\begin{tabular}{>{\centering\arraybackslash}m{1.5cm}cccm{6cm}}
\Xhline{1pt}
\textbf{Dataset} & \textbf{Size} & \textbf{\# Features} & \textbf{ML Problem}\\
\hline
\hline
Bike & 17379 & 13 & Predicting count of total rental bikes including both casual and registered.\\
\hline
Concrete & 1030 & 8 & Predicting concrete compressive strength.\\
\hline
Parkinsons & 5875 & 19 & Predicting clinician's total UPDRS score, linearly interpolated.\\
\hline
Realstate & 414 & 6 &  Predicting house price of unit area.\\
\Xhline{1pt}
\end{tabular}
\endgroup
\caption{\label{dataset_description} This table provides descriptions of the datasets used for experiments of feature engineering. These datasets are intended for solving regression problems.}
\end{table*}

\clearpage



\end{document}